\documentclass[11pt]{article}
\usepackage{naaclhlt2009}
\usepackage{times}
\makeatletter
\newcommand{\@BIBLABEL}{\@emptybiblabel}
\newcommand{\@emptybiblabel}[1]{}
\makeatother
\usepackage{latexsym}
\usepackage{epsf}
\usepackage{graphicx}
\usepackage{hyperref} 
\title{Taking into Account the Differences between Actively and Passively Acquired Data: 
The Case of Active Learning with Support Vector Machines for Imbalanced Datasets}

\author{
  Michael Bloodgood\Thanks{ This research was conducted while the first author was a PhD student at the University of Delaware.} \\
  Human Language Technology \\
  Center of Excellence \\
  Johns Hopkins University \\
  Baltimore, MD 21211 USA \\
  {\tt bloodgood@jhu.edu} \And
  K. Vijay-Shanker \\
  Computer and Information \\
  Sciences Department \\
  University of Delaware \\
  Newark, DE 19716 USA \\
  {\tt vijay@cis.udel.edu}}

\date{}

\begin{document}
\maketitle
\begin{abstract}
Actively sampled data can have very different characteristics than passively sampled data. 
Therefore, it's promising to investigate using different inference procedures during AL 
than are used during passive learning (PL). 
This general idea is explored in detail for the focused case of AL with cost-weighted SVMs for imbalanced data, a 
situation that arises for many HLT tasks. The key idea behind the proposed InitPA method for addressing imbalance is to 
base cost models during AL on an estimate of overall corpus imbalance computed via a small unbiased sample rather than 
the imbalance in the labeled training data, which is the leading method used during PL. 
\end{abstract}

\section{Introduction} \label{intro}

Recently there has been considerable interest in using active learning (AL) to reduce HLT annotation burdens. 
Actively sampled data can have different characteristics than passively sampled data and therefore, this
paper proposes modifying algorithms used to infer models during AL.  Since most AL research 
assumes the same learning algorithms will be used
during AL as during passive learning\footnote{Passive learning refers to the typical supervised learning setup where the learner does
not actively select its training data.} (PL), this paper opens up a new thread of AL research that accounts for the differences between 
passively and actively sampled data. 

The specific case focused on in this paper is that of AL with SVMs (AL-SVM)
for imbalanced datasets\footnote{This paper focuses on the fundamental case of binary classification where class imbalance
arises because the positive examples are rarer than the negative examples, a situation that naturally arises for many HLT
tasks.}. Collectively, the factors: interest in AL, 
widespread class imbalance for many HLT tasks, interest in using SVMs, 
and PL research showing that SVM performance can be improved substantially by addressing imbalance, indicate
the importance of the case of AL with SVMs with imbalanced data. 

Extensive PL research has shown that learning algorithms' performance degrades for imbalanced datasets and
techniques have been developed that prevent this degradation. However, 
to date, relatively little work has addressed imbalance during AL (see Section~\ref{related}). 
In contrast to previous work, this paper advocates that the AL scenario brings out the need to modify PL
approaches to dealing with imbalance. In particular, a new method is developed for cost-weighted SVMs 
that estimates a cost model based on overall corpus imbalance rather than the imbalance in the so far 
labeled training data. Section~\ref{related} discusses related work, Section~\ref{setup} discusses the
experimental setup, Section~\ref{methods} presents the new method called InitPA, 
Section~\ref{eval} evaluates InitPA, and Section~\ref{conclusions} concludes. 

\section{Related Work} \label{related}

A problem with imbalanced data is that the class boundary (hyperplane) learned by SVMs can be too close 
to the positive (pos) examples and then recall suffers. Many approaches have been presented for overcoming this
problem {\em in the PL setting}. Many require substantially longer training times or extra training data to tune parameters 
and thus are not ideal for use during AL. Cost-weighted SVMs (cwSVMs), on the other hand, {\em are} a promising
approach for use with AL: they impose no extra training overhead. cwSVMs introduce unequal cost factors so
the optimization problem solved becomes:

Minimize:
\vspace*{-.1cm} 
\begin{equation}
\frac{1}{2}\|\vec{w}\|^{2} + C_{+}\sum_{i:y_{i}=+1}\xi_{i} + C_{-}\sum_{i:y_{i}=-1}\xi_{i}
\vspace*{-.4cm}
\end{equation}

Subject to: 
\vspace*{-.2cm}
\begin{equation}
\forall k : y_k\left[\vec{w} \cdot \vec{x}_k + b \right] \ge 1 - \xi_{k},
%\vspace*{-.2cm}
\end{equation}
where $(\vec{w},b)$ represents the learned hyperplane, $\vec{x}_k$ is the feature vector for example $k$, 
$y_k$ is the label for example $k$, $\xi_k = max(0,1-y_k(\vec{w}_k \cdot \vec{x}_k + b))$ is the slack
variable for example $k$, and $C_+$ and $C_-$ are user-defined cost factors. 

The most important part for this paper are the cost factors $C_{+}$ and $C_{-}$. The ratio
$\frac{C_{+}}{C_{-}}$ quantifies the importance of reducing slack error on pos train examples relative to
reducing slack error on negative (neg) train examples. The value of the ratio is crucial for balancing the precision
recall tradeoff well. \cite{morik1999} showed that during PL, setting 
$\frac{C_{+}}{C_{-}} = \frac{\mbox{\# of neg training examples}}{\mbox{\# of pos training examples}}$ is an effective
heuristic. Section~\ref{methods} explores using this heuristic {\em during AL} and explains a modified
heuristic that could work better during AL. 

\cite{ertekin2007b} propose using the balancing of training data that occurs as a result of AL-SVM to
handle imbalance and do not use any further measures to address imbalance. 
\cite{zhu2007} used resampling to address imbalance and based the amount of resampling, which is the analog of our cost model, 
on the amount of imbalance
in the current set of labeled train data, as PL approaches do. 
In contrast, the InitPA approach in Section~\ref{methods} 
bases its cost models on overall (unlabeled) corpus imbalance rather than the amount of imbalance in the
current set of labeled data. 

\section{Experimental Setup} \label{setup}

We use relation extraction (RE) and text classification (TC) datasets and
SVM$^{light}$ \cite{joachims1999} for training the SVMs. For RE, we use AImed, previously used to train 
protein interaction extraction systems (\cite{giuliano2006}). As in previous work, we cast RE as a binary classification
task (14.94\% of the examples in AImed are positive). We use the $K_{GC}$ kernel from \cite{giuliano2006}, one of the 
highest-performing systems on AImed to date and perform 10-fold cross validation. For TC, we use the Reuters-21578
ModApte split. Since a document may belong to more than one category, each
category is treated as a separate binary classification problem, as in \cite{joachims1998}. As in \cite{joachims1998}, we
use the ten largest categories, which have imbalances ranging from 1.88\% to 29.96\%. 

\section{AL-SVM Methods for Addressing Class Imbalance} \label{methods}

\begin{figure}
\begin{center}
\includegraphics[width=8cm,height=4cm]{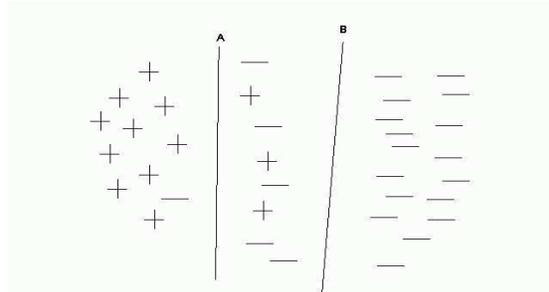}
\vspace{-1cm}
\caption{\label{f:hyperplanes}Hyperplane B was trained with a higher $\frac{C_+}{C_-}$ ratio than hyperplane A was trained with.}
\end{center}
\end{figure}

The key question when using cwSVMs is how to set the ratio $\frac{C_{+}}{C_{-}}$. 
Increasing it will typically shift the learned hyperplane so recall is increased and precision is
decreased (see Figure~\ref{f:hyperplanes} for a hypothetical example). 
Let PA$=\frac{C_+}{C_-}$.\footnote{PA stands for positive amplification and gives us a concise 
way to denote the fraction $\frac{C_+}{C_-}$, which doesn't have a standard name.}
How should the PA be set during AL-SVM?

We propose two approaches: one sets the PA based on the level of imbalance in the labeled training data and
one aims to set the PA based on an estimate of overall corpus imbalance, 
{\em which can drastically differ from the level of imbalance in actively sampled training data}.
The first method is called CurrentPA, depicted in Figure~\ref{f:CurrentPA}. 
Note that in step 0 of the loop, PA is set based on the distribution of positive and negative examples 
in the {\em current} set of labeled data. 
However, observe that during AL the 
ratio $\frac{\mbox{\# neg labeled examples}}
{\mbox{\# pos labeled examples}}$ in the current set of labeled data gets skewed from the ratio in the entire corpus
because AL systematically selects the examples that are closest to the current model's hyperplane and this tends
to select more positive examples than random selection would select (see also \cite{ertekin2007b}).
% CurrentPA Figure
\begin{figure}
\begin{center}
\begin{tabbing}
{\bf Input:}\=\\
\>$L =$ small initial set of labeled data\\
\>$U =$ large pool of unlabeled data\\
{\bf Loop} until stopping criterion is met:\\
\>0. Set PA $= \frac{|\{x \in Labeled : f(x) = -1\}|}{|\{x \in L : f(x) = +1\}|}$ \\
\>where $f$ is the function we desire to learn. \\
\>1. Train an SVM with $C_+$ and $C_-$ set such \\
\>that $\frac{C_+}{C_-} =$ PA and obtain
hyperplane $h \; .$\protect\footnotemark\\
\>2. $batch \leftarrow$ select k points from $U$ that are\\ 
\>closest to $h$ and request their labels.\protect\footnotemark\\ 
\>3. $U = U - batch \; .$\\
\>4. $L = L \cup batch \; .$\\
{\bf End Loop}\\
\end{tabbing}
\end{center}
\vspace*{-.5cm}
\caption{\label{f:CurrentPA} The CurrentPA algorithm}
\end{figure}
\addtocounter{footnote}{-1}\footnotetext{We use SVM$^{light}$'s default value for $C_-$.}
\addtocounter{footnote}{1}\footnotetext{In our experiments, batch size is 20.}

\begin{figure}
\begin{center}
\includegraphics[width=8cm,height=5cm]{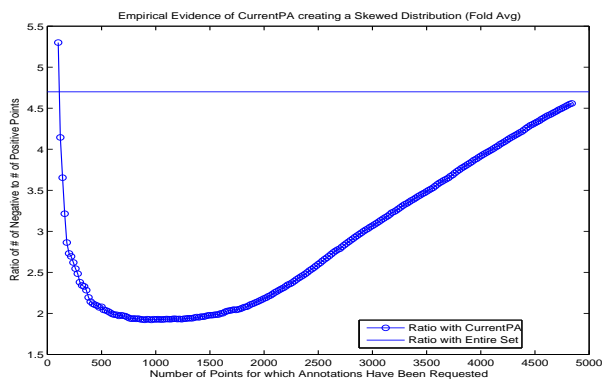}
\caption{\label{f:aimedSkew}Illustration of AL skewing the distribution of pos/neg points on AImed.}
\end{center}
\end{figure}

Empirical evidence of this distribution skew is illustrated in Figure~\ref{f:aimedSkew}.
The trend toward balanced datasets during AL could mislead and cause us to underestimate the PA. 

Therefore, our next algorithm aims to set the PA based on the ratio of neg to pos instances in the
entire corpus. However, since we don't have labels for the entire corpus, we don't know this ratio. But by  
using a small
initial sample of labeled data, we can estimate this ratio with high confidence. This estimate can then be used for
setting the PA throughout the AL process. We call this method of setting the PA based on a small initial set of labeled
data the InitPA method. It is like CurrentPA except we move {\em Step 0} to be executed one time before the loop and then use
that same PA value on each iteration of the AL loop.

To guide what size to make the initial set of labeled data, one
can determine the sample size required to estimate the proportion of positives in a finite population to within
sampling error $e$ with a desired level of confidence using standard statistical techniques found in many
college-level statistics references such as \cite{berenson}. For example, carrying out the 
computations on the AImed dataset shows that a size of 100 enables us to 
be 95\% confident that our proportion estimate is within 0.0739 of the true proportion. 
In our experiments, we used an initial labeled set of size 100. 

\section{Evaluation} \label{eval}

\begin{figure}[t]
\begin{center}
\includegraphics[width=8cm,height=5cm]{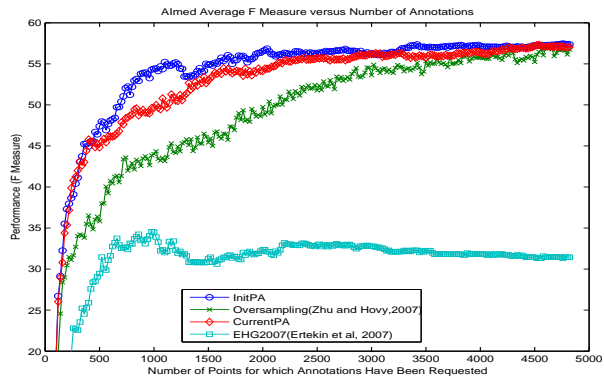}
\caption{\label{f:aimedCurves}AImed learning curves. y-axis is from 20\% to 60\%.}
\end{center}
\end{figure}

\begin{figure}[t]
\begin{center}
\includegraphics[width=8cm,height=5cm]{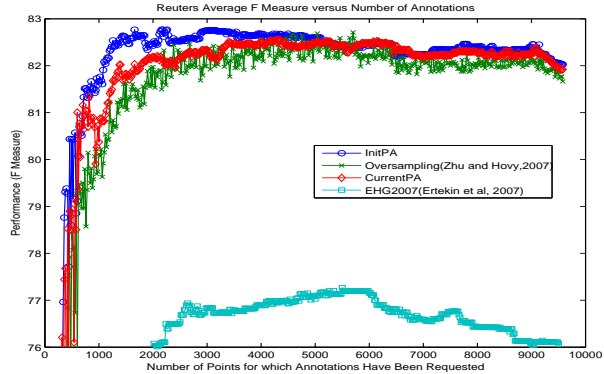}
\caption{\label{f:reutersCurves}Reuters learning curves. y-axis is from 76\% to 83\%.}
\end{center}
\end{figure}

In addition to InitPA and CurrentPA, we also implemented the methods from \cite{ertekin2007b,zhu2007}.
We implemented oversampling by duplicating points and by BootOS \cite{zhu2007}. To 
avoid cluttering the graphs, we only show the highest-performing oversampling variant, which was by duplicating points.
Learning curves are presented in Figures \ref{f:aimedCurves} and \ref{f:reutersCurves}.

Note InitPA is the highest-performing method 
for all datasets, especially in the practically important area of where the learning curves begin to plateau. 
This area is important because this is around where we would want to stop AL \cite{bloodgood2009b}.

Observe that the gains of InitPA over CurrentPA are smaller for Reuters. For some 
Reuters categories, InitPA and CurrentPA have nearly identical performance. Applying the models learned by
CurrentPA at each round of AL on the data used to train the model reveals that the recall on the training data is
nearly 100\% for those categories where InitPA/CurrentPA perform similarly. Increasing the relative penalty for slack error
on positive training points will not have much impact if (nearly) all of the pos train points are already classified correctly.
Thus, in situations where models are already achieving nearly 100\% recall on their train data, InitPA is not expected to
outperform CurrentPA. 

The hyperplanes learned during AL-SVM serve two purposes: {\em sampling} - they govern which unlabeled points will be selected for
human annotation, and {\em predicting} - when AL stops, the most recently learned hyperplane is used for classifying test
data. 
Although all AL-SVM approaches we're aware of use the same hyperplane at each round of AL for both of these purposes, this
is not required. We compared InitPA with hybrid approaches where hyperplanes trained using an InitPA cost model are used for
sampling and hyperplanes trained using a CurrentPA cost model are used for predicting, and vice-versa, and found that InitPA
performed better than both of these hybrid approaches. This indicates that the InitPA cost model yields hyperplanes that are
better for both sampling and predicting. 

\section{Conclusions} \label{conclusions}

We've made the case for the importance of AL-SVM for imbalanced datasets and showed that the AL scenario calls for 
modifications to PL approaches to addressing imbalance. 
For AL-SVM, the key idea behind InitPA is to base cost models on an estimate of overall corpus imbalance rather than 
the class imbalance in the so far labeled data. 
The practical utility of the InitPA method was demonstrated empirically; situations where InitPA won't help that much 
were made clear; and analysis showed that the sources of InitPA's gains were from both better sampling and
better predictive models. 

InitPA is an instantiation of a more general idea of {\em not} using the same inference algorithms during AL as 
during PL but instead modifying inference algorithms to suit esoteric characteristics of actively sampled data. 
This is an idea that has seen relatively little exploration and is ripe for further investigation. 

\bibliographystyle{naaclhlt2009}
\bibliography{paper}

\begin{thebibliography}{}

\bibitem[\protect\citename{Berenson \bgroup et al.\egroup }1988]{berenson}
Mark~L. Berenson, David~M. Levine, and David Rindskopf.
\newblock 1988.
\newblock {\em Applied Statistics}.
\newblock Prentice-Hall, Englewood Cliffs, NJ.

\bibitem[\protect\citename{Bloodgood and Vijay-Shanker}2009]{bloodgood2009b}
Michael Bloodgood and K.~Vijay-Shanker.
\newblock 2009.
\newblock A method for stopping active learning based on stabilizing
  predictions and the need for user-adjustable stopping.
\newblock In {\em CoNLL}.

\bibitem[\protect\citename{Ertekin \bgroup et al.\egroup }2007]{ertekin2007b}
Seyda Ertekin, Jian Huang, L{\'e}on Bottou, and C.~Lee Giles.
\newblock 2007.
\newblock Learning on the border: active learning in imbalanced data
  classification.
\newblock In {\em CIKM}.

\bibitem[\protect\citename{Giuliano \bgroup et al.\egroup }2006]{giuliano2006}
Claudio Giuliano, Alberto Lavelli, and Lorenza Romano.
\newblock 2006.
\newblock Exploiting shallow linguistic information for relation extraction
  from biomedical literature.
\newblock In {\em EACL}.

\bibitem[\protect\citename{Joachims}1998]{joachims1998}
Thorsten Joachims.
\newblock 1998.
\newblock Text categorization with suport vector machines: Learning with many
  relevant features.
\newblock In {\em ECML}, pages 137--142.

\bibitem[\protect\citename{Joachims}1999]{joachims1999}
Thorsten Joachims.
\newblock 1999.
\newblock Making large-scale {SVM} learning practical.
\newblock In {\em Advances in Kernel Methods -- Support Vector Learning}, pages
  169--184.

\bibitem[\protect\citename{Morik \bgroup et al.\egroup }1999]{morik1999}
Katharina Morik, Peter Brockhausen, and Thorsten Joachims.
\newblock 1999.
\newblock Combining statistical learning with a knowledge-based approach - a
  case study in intensive care monitoring.
\newblock In {\em ICML}, pages 268--277.

\bibitem[\protect\citename{Zhu and Hovy}2007]{zhu2007}
Jingbo Zhu and Eduard Hovy.
\newblock 2007.
\newblock Active learning for word sense disambiguation with methods for
  addressing the class imbalance problem.
\newblock In {\em EMNLP-CoNLL}.

\end{thebibliography}

\end{document}